\documentclass[conference]{IEEEtran}
\IEEEoverridecommandlockouts
\usepackage{cite}
\usepackage{amsmath,amssymb,amsfonts}
\usepackage{algorithmic}
\usepackage{graphicx}
\usepackage{textcomp}
\usepackage{xcolor}
\usepackage{placeins}

\def\BibTeX{{\rm B\kern-.05em{\sc i\kern-.025em b}\kern-.08em
    T\kern-.1667em\lower.7ex\hbox{E}\kern-.125emX}}

\makeatletter 
\newcommand{\linebreakand}{%
  \end{@IEEEauthorhalign}
  \hfill\mbox{}\par
  \mbox{}\hfill\begin{@IEEEauthorhalign}
}
\makeatother 
\begin{document}

\title{Enhancing Document-Level Question Answering via Multi-Hop Retrieval-Augmented Generation with LLaMA 3\\}

\author{
\IEEEauthorblockN{Xinyue Huang*}
\IEEEauthorblockA{\textit{Independent Researcher} \\
New York, USA \\
huangxinyue8616@gmail.com}
\and
\IEEEauthorblockN{Ziqi Lin}
\IEEEauthorblockA{\textit{Cornell University } \\
New Jersey, USA \\
zl825@cornell.edu}
\and
\IEEEauthorblockN{Fang Sun}
\IEEEauthorblockA{\textit{University of Southern California} \\
Los Angeles, USA \\
fangsun@usc.edu}
\and
\linebreakand
\IEEEauthorblockN{Wenchao Zhang}
\IEEEauthorblockA{\textit{Independent Researcher} \\
New Jersey, USA \\
wenchao.zhang@rutgers.edu}
\and
\IEEEauthorblockN{Kejian Tong}
\IEEEauthorblockA{\textit{Independent Researcher} \\
Mukilteo, USA \\
tongcs2021@gmail.com}
\and
\IEEEauthorblockN{Yunbo Liu}
\IEEEauthorblockA{\textit{Independent Researcher} \\
New York, USA \\
chrisliu38387@gmail.com}
}

\maketitle

\begin{abstract}
This paper presents a novel Retrieval-Augmented Generation (RAG) framework tailored for complex question answering tasks, addressing challenges in multi-hop reasoning and contextual understanding across lengthy documents. Built upon LLaMA 3, the framework integrates a dense retrieval module with advanced context fusion and multi-hop reasoning mechanisms, enabling more accurate and coherent response generation. A joint optimization strategy combining retrieval likelihood and generation cross-entropy improves the model’s robustness and adaptability. Experimental results show that the proposed system outperforms existing retrieval-augmented and generative baselines, confirming its effectiveness in delivering precise, contextually grounded answers.

\end{abstract}


\section{Introduction}
Understanding complex question answering (QA) tasks requires deep comprehension of documents containing numbers, legal texts, and intricate language. Large language models (LLMs) often struggle to effectively retrieve and reason over dispersed pieces of information. Retrieval-Augmented Generation (RAG), which integrates retrieval and generation, has shown promising results. However, many existing RAG models still face limitations in multi-hop reasoning and context fusion, which are crucial for tasks involving linked reports, statements, and structured content. Recent advances have addressed these challenges in part—for instance, Dai et al.\cite{dai2025cab} employed contrastive augmentation to strengthen retrieval,  Wang et al.\cite{wang2025attention} introduced an attention-based architecture for improved context comprehension.

In this study, we propose a multi-module RAG framework built on LLaMA 3 with enhanced retrieval and reasoning capabilities. The system incorporates a query-document embedding module that generates high-dimensional representations and retrieves relevant content from a vector database. To overcome single-hop limitations, we introduce a multi-hop reasoning module that incrementally aggregates context across documents via attention mechanisms. A joint optimization strategy combining retrieval likelihood and generation cross-entropy further improves both retrieval precision and generation quality. Overall, the framework demonstrates improved performance in answering complex queries requiring deep contextual understanding. 

Beyond improving general document QA, our methodology also supports high-stakes domains—fraud investigation, regulatory compliance and risk analysis—by enabling accurate multi-document retrieval and reasoning over lengthy, cross-referenced records (e.g., suspicious activity reports, customer disclosures and transaction logs). This capability facilitates automated early fraud detection, streamlined compliance workflows and enhanced transparency in financial operations—key priorities for institutions and regulators.
Building on these findings, FinLLaMA-RAG holds significant potential in tax compliance and strategy through two key applications. First, it can empower individual taxpayers and small businesses by serving as a virtual tax assistant. Leveraging its multi-hop retrieval and reasoning, the system can dynamically retrieve relevant sections of tax code and official publications and fuse context (e.g. income type, filing status) to provide personalized, legally accurate guidance on deductions, credits, and filing requirements—helping users maximize benefits and avoid errors that often lead to inquiries or penalties.

\section{Related Work}
Choi et al.\cite{choi2025finder} made FinDER, a dataset for financial QA and RAG tests, to solve the lack of good financial data. Kim et al.\cite{kim2025optimizing} improved retrieval for financial QA by adding a multi-stage optimization that raises document relevance, but their work focuses more on retrieval than text generation. Chen et al.\cite{chen2024coarse} created a coarse-to-fine 3D reconstruction system with transformers. While it works in vision tasks, it shows how attention can help in text retrieval too. Guan et al.\cite{guan2025breast} used machine learning to predict breast cancer with network analysis, giving ideas about modeling complex links, though in a medical setting. Luo, Wang, and Guo \cite{luo2025gemini} introduce Gemini-GraphQA, a graph question answering framework that integrates the Gemini large language model with a graph neural network encoder, a graph solver network to translate natural language into executable graph code, and a retrieval-augmented generation module—enhanced by an execution correctness loss—to ensure syntactic and functional accuracy, achieving state-of-the-art performance on diverse graph reasoning tasks.

Chen et al.\cite{chen2024fintextqa} made FinTextQA, a dataset for long-form financial QA, which helps with large-context understanding but does not add new RAG methods. Sarmah et al.\cite{sarmah2024hybridrag} proposed HybridRAG, which mixes knowledge graphs with vector retrieval to improve information extraction, but its multi-hop part is still simple. Iaroshev et al.\cite{iaroshev2024evaluating} tested RAG systems on financial reports and showed that challenges remain in dealing with detailed domain language and links between documents.Yu \cite{yu2025towards} introduces DynaSched-Net, a dual-network framework that combines a Deep Q-Network–based reinforcement learning scheduler with a hybrid LSTM-Transformer workload predictor—optimized via a joint loss function and stabilized by experience replay and target network updates—to enable real-time adaptive cloud resource scheduling that outperforms traditional FCFS and RR methods. Their results also pointed out that current systems often fail when financial questions need reasoning over multiple sections, which shows a need for better ways to combine retrieved data into a full answer. 

Lin et al.\cite{lin2025tax} propose a vector‐weighted average algorithm–optimized kernel Extreme Learning Machine for national tax revenue ratio prediction, achieving R² values of 0.995 (training) and 0.994 (test) with RMSEs of 0.185 and 0.177, respectively, demonstrating excellent generalization and stability for tax forecasting. In many cases, the retrieved documents are relevant but the generated answers miss key context, which limits the system’s real use. This makes it clear that a better model should focus on both improving retrieval precision and making sure the generation part fully uses all the retrieved information. Guo and Yu \cite{guo2025privacypreservenet} propose PrivacyPreserveNet, a novel multilevel privacy-preserving framework for multimodal large language models that integrates differential privacy-enhanced pretraining, privacy-aware gradient clipping, and noise-injected attention mechanisms to safeguard sensitive text, image, and audio data without sacrificing task performance. 

\section{Methodology}
\begin{figure}[htbp]
  \centering
  \includegraphics[width=0.5\textwidth]{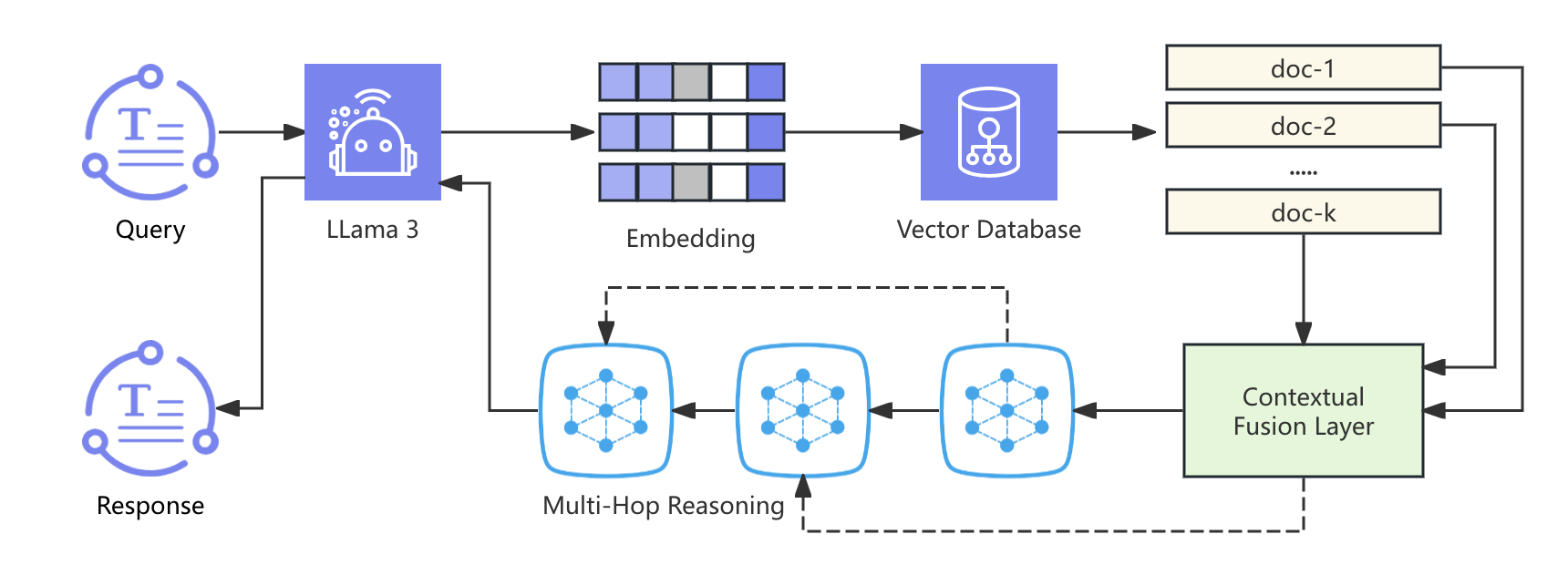}
  \caption{The FinLLaMA-RAG base on LLaMA 3 using multi-hop reasoning module}
  \label{fig:model}
\end{figure}

In this section, we introduce FinLLaMA-RAG, an advanced Retrieval-Augmented Generation (RAG) model designed for document analysis. Leveraging the LLaMA 3 model, FinLLaMA-RAG integrates a multi-hop reasoning module to traverse complex data, enhancing the accuracy and relevance of generated responses. The system employs a contextual fusion layer to aggregate information from multiple document chunks, facilitating comprehensive understanding. A novel loss function balances retrieval accuracy and generation quality, optimizing both components simultaneously. Experimental evaluations demonstrate that FinLLaMA-RAG outperforms existing models in handling intricate queries, offering a robust solution for document analysis. The pipeline of our approach is shown in Fig.~\ref{fig:model}.

\subsection{Query Embedding Module}
The input query \(q\) is transformed into a dense vector representation \(\mathbf{q}\) using a pre-trained LLaMA 3 model:
\begin{equation}
\mathbf{q} \;=\;\mathrm{LLaMA3}_{\mathrm{embed}}(q)
\label{eq:query_embed}
\end{equation}
This embedding captures the semantic meaning of the query, facilitating efficient retrieval of relevant document chunks.

\subsection{Document Retrieval Module}
Utilizing the query embedding \(\mathbf{q}\), the system retrieves the top-\(k\) most relevant document chunks \(\{d_1, d_2, \dots, d_k\}\) from a vector database. The relevance of each chunk \(d_i\) is assessed using cosine similarity:
\begin{equation}
\mathrm{sim}(q, d_i) \;=\;\frac{\mathbf{q}^\top \mathbf{d}_i}{\|\mathbf{q}\|\;\|\mathbf{d}_i\|}
\label{eq:cosine_sim}
\end{equation}
where \(\mathbf{d}_i\) is the embedding of chunk \(d_i\).

\subsection{Contextual Fusion Layer}
To enhance the representation of the retrieved chunks, a contextual fusion layer aggregates the embeddings:
\begin{equation}
\mathbf{D}_{\mathrm{agg}} \;=\;\sum_{i=1}^{k} \alpha_i \,\mathbf{d}_i
\label{eq:fusion}
\end{equation}
with attention weights
\begin{equation}
\alpha_i \;=\;\frac{\exp\!\bigl(\mathrm{sim}(q, d_i)\bigr)}{\sum_{j=1}^{k}\exp\!\bigl(\mathrm{sim}(q, d_j)\bigr)}.
\label{eq:attention_weight}
\end{equation}

\subsection{Multi-Hop Reasoning Module}
\begin{figure*}[htbp]
  \centering
  \includegraphics[width=\textwidth]{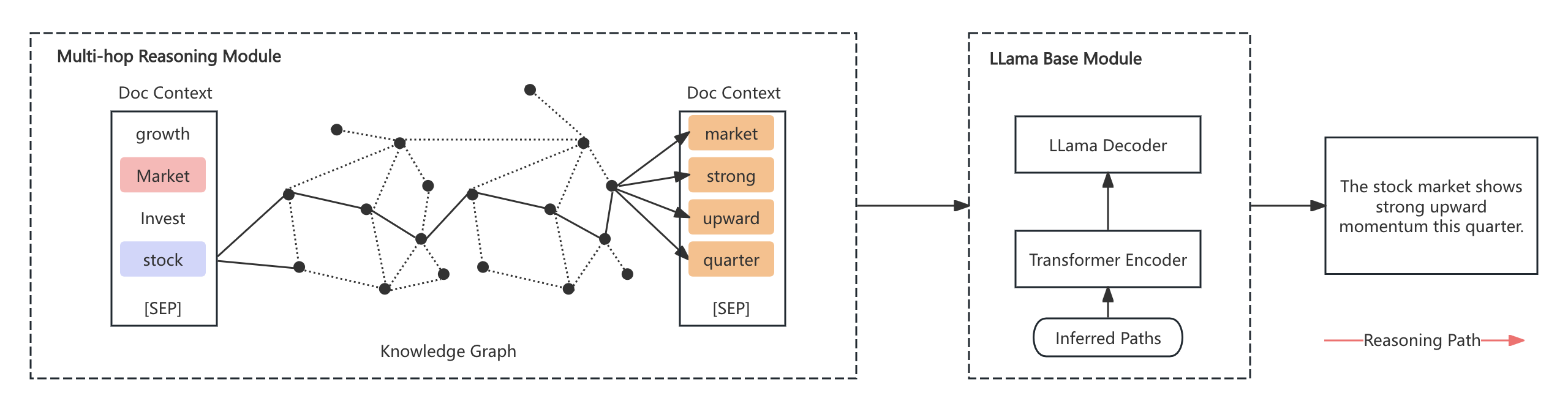}
  \caption{The pipeline of the Multi-Hop Reasoning Module.}
  \label{fig:model2}
\end{figure*}

The multi-hop reasoning module performs iterative updates over the aggregated representation:
\begin{equation}
\mathbf{D}_{\mathrm{hop}}^{(t)} \;=\;\mathrm{LLaMA3}_{\mathrm{hop}}\bigl(\mathbf{D}_{\mathrm{hop}}^{(t-1)},\,\mathbf{q}\bigr),
\quad
\mathbf{D}_{\mathrm{hop}}^{(0)} = \mathbf{D}_{\mathrm{agg}},
\label{eq:multi_hop}
\end{equation}
for \(t=1,\dots,T\). The pipeline of this module is shown in Fig.~\ref{fig:model2}.

\subsection{Generation Module}
The final representation \(\mathbf{D}_{\mathrm{hop}}^{(T)}\) is passed to the LLaMA 3-based generation module, which produces the response \(r\) to the input query \(q\):
\begin{equation}
r \;=\;\mathrm{LLaMA3}_{\mathrm{gen}}\bigl(\mathbf{D}_{\mathrm{hop}}^{(T)},\,q\bigr).
\label{eq:generation}
\end{equation}

\subsection{Loss Function}
\begin{figure}[htbp]
  \centering
  \includegraphics[width=0.4\textwidth]{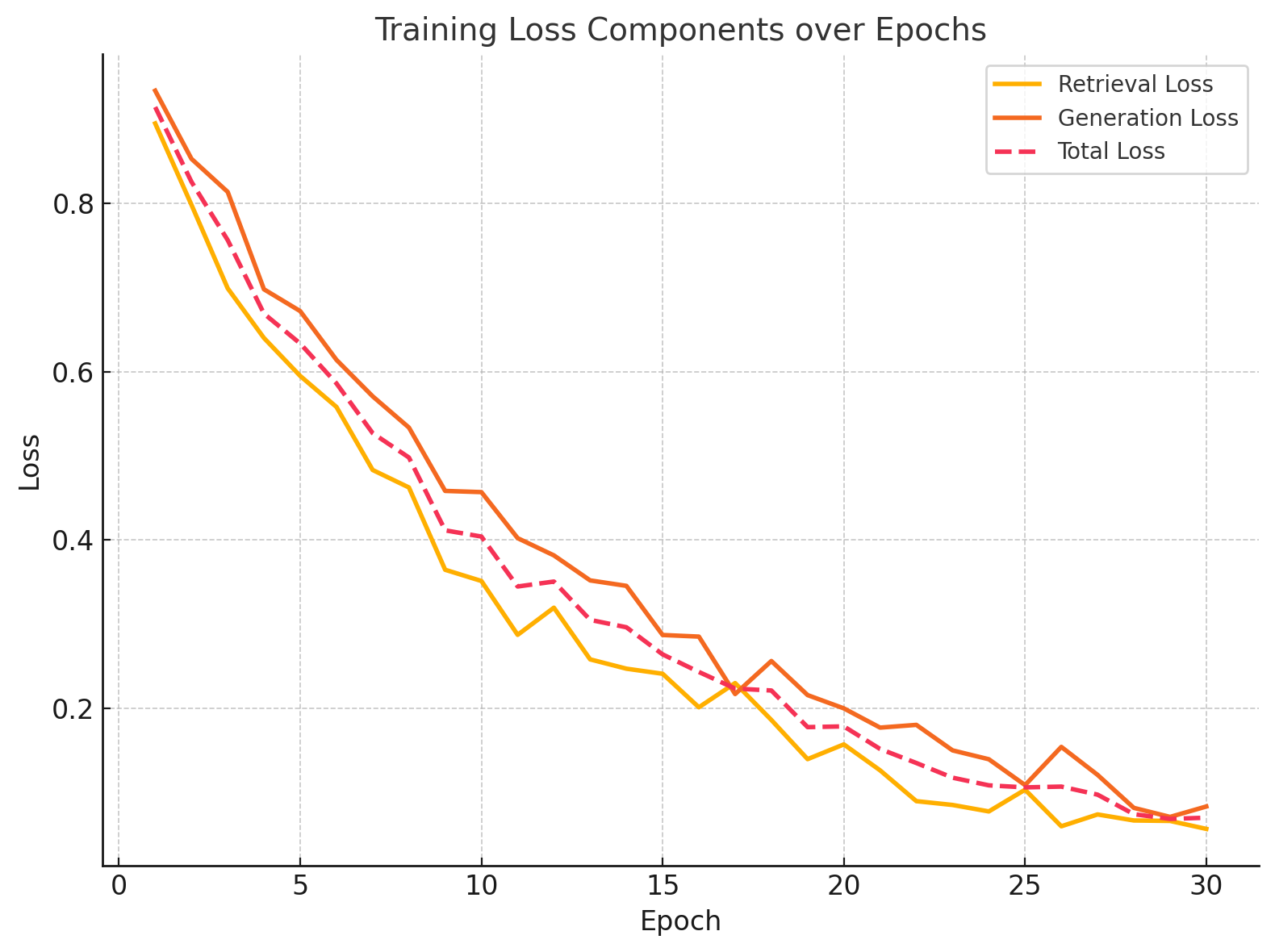}
  \caption{Training loss components over epochs: retrieval loss, generation loss, and total loss}
  \label{fig:loss}
\end{figure}

The training objective combines retrieval accuracy and generation quality. The retrieval loss is
\begin{equation}
L_{\mathrm{retrieval}} \;=\; -\log \frac{\exp\!\bigl(\mathrm{sim}(q, d_{\mathrm{true}})\bigr)}{\sum_{i=1}^{k}\exp\!\bigl(\mathrm{sim}(q, d_i)\bigr)},
\label{eq:retrieval_loss}
\end{equation}
and the generation loss is
\begin{equation}
L_{\mathrm{generation}} \;=\; -\sum_{t=1}^{T} \log P\bigl(r_t \mid r_{<t}, \mathbf{D}_{\mathrm{hop}}^{(T)}, q\bigr).
\label{eq:generation_loss}
\end{equation}
The total loss is a weighted sum:
\begin{equation}
L_{\mathrm{total}} \;=\; \lambda_{\mathrm{retrieval}}\,L_{\mathrm{retrieval}}
+\;\lambda_{\mathrm{generation}}\,L_{\mathrm{generation}},
\label{eq:total_loss}
\end{equation}
where \(\lambda_{\mathrm{retrieval}}\) and \(\lambda_{\mathrm{generation}}\) are hyperparameters. Training loss curves are shown in Fig.~\ref{fig:loss}.

\subsection{Integration of Large-Scale Document Embeddings}
One key innovation of FinLLaMA-RAG is the integration of large-scale pre-trained models like LLaMA 3 with efficient document retrieval and re-ranking mechanisms. By embedding both the query and chunks into high-dimensional vectors and applying similarity-based retrieval, the model can efficiently handle vast collections of documents. Combining retrieval-augmented information with the generative capabilities of LLaMA 3 enables more accurate, contextually relevant responses.
FinLLaMA-RAG can streamline international tax strategy for multinational corporations. By parsing and comparing complex regulations—such as bilateral treaties and global tax frameworks—it can rapidly benchmark transfer-pricing policies across jurisdictions. This reduces research time, enhances accuracy of intercompany pricing, and generates an audit-ready trail of citations, supporting both corporate documentation and regulatory oversight to minimize costly disputes.

\subsection{Multi-Hop Reasoning Across Hierarchical Data}
\begin{figure}[htbp]
  \centering
  \includegraphics[width=0.5\textwidth]{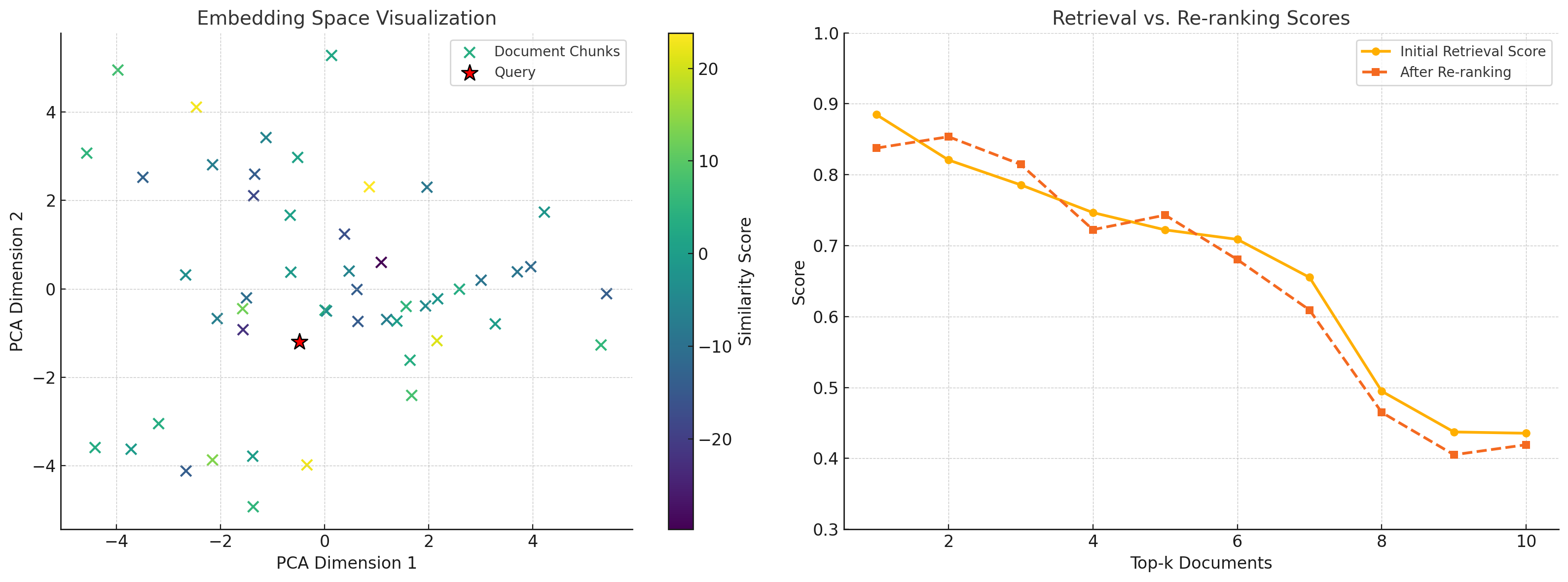}
  \caption{(Left) Visualization of query and document embeddings in 2D via PCA. (Right) Comparison of initial retrieval scores and re-ranked scores across top-\(k\) documents.}
  \label{fig:embedding_integration}
\end{figure}

Another innovation is the use of the multi-hop reasoning module, which enables iterative reasoning across multiple document sections. This approach allows for a more comprehensive understanding of information, as the model can reason over interconnected sections to extract insights. This is especially crucial in analysis scenarios where a question may require synthesizing information from several document parts. As shown in Fig.~\ref{fig:embedding_integration}, the left panel visualizes the embedding space via PCA, and the right panel compares initial retrieval scores with re-ranked scores.

\subsection{Data Preprocessing}

Raw documents \(d_{\mathrm{raw}}\) are cleaned by
\begin{equation}
d_{\mathrm{clean}} = \mathrm{Clean}\bigl(d_{\mathrm{raw}}\bigr)
\label{eq:clean}
\end{equation}

The cleaned text is tokenized into IDs:
\begin{equation}
d_{\mathrm{tok}} = \bigl[\mathrm{ID}(t_1),\dots,\mathrm{ID}(t_n)\bigr]
\label{eq:tokenize}
\end{equation}

Embeddings are generated and indexed for retrieval:
\begin{align}
\mathbf{q} &= \mathrm{LLaMA3}_{\mathrm{embed}}(q)
\label{eq:embed_q}\\
\mathbf{d}_i &= \mathrm{LLaMA3}_{\mathrm{embed}}(d_i)
\label{eq:embed_d}\\
\mathrm{sim}(q,d_i) &= \frac{\mathbf{q}^\top \mathbf{d}_i}{\|\mathbf{q}\|\|\mathbf{d}_i\|}
\label{eq:cos_sim}
\end{align}

\section{Evaluation Metrics}

The model performance is evaluated using several key metrics:

\subsection{nDCG@10}
nDCG@10 evaluates the ranking of the top-10 retrieved documents. It is calculated as:
\begin{equation}
\text{nDCG@10} = \frac{1}{Z} \sum_{i=1}^{10} \frac{2^{\text{rel}_i} - 1}{\log_2(i + 1)}
\end{equation}

\subsection{BLEU}
BLEU measures the overlap of n-grams between the predicted and reference responses. It is computed as:
\begin{equation}
\text{BLEU} = \exp \left( \frac{1}{N} \sum_{n=1}^{N} \log p_n \right)
\end{equation}

\subsection{ROUGE-L}
ROUGE-L measures the longest common subsequence (LCS) between predicted and reference responses:
\begin{equation}
\text{ROUGE-L} = \frac{LCS(\text{reference}, \text{prediction})}{\text{length of reference}}
\end{equation}

\subsection{F1 Score}
The F1 score is calculated as:
\begin{equation}
\text{F1} = 2 \times \frac{\text{precision} \times \text{recall}}{\text{precision} + \text{recall}}
\end{equation}

\section{Experiment Results}

\begin{figure}[htbp]
  \centering
  \includegraphics[width=0.5\textwidth]{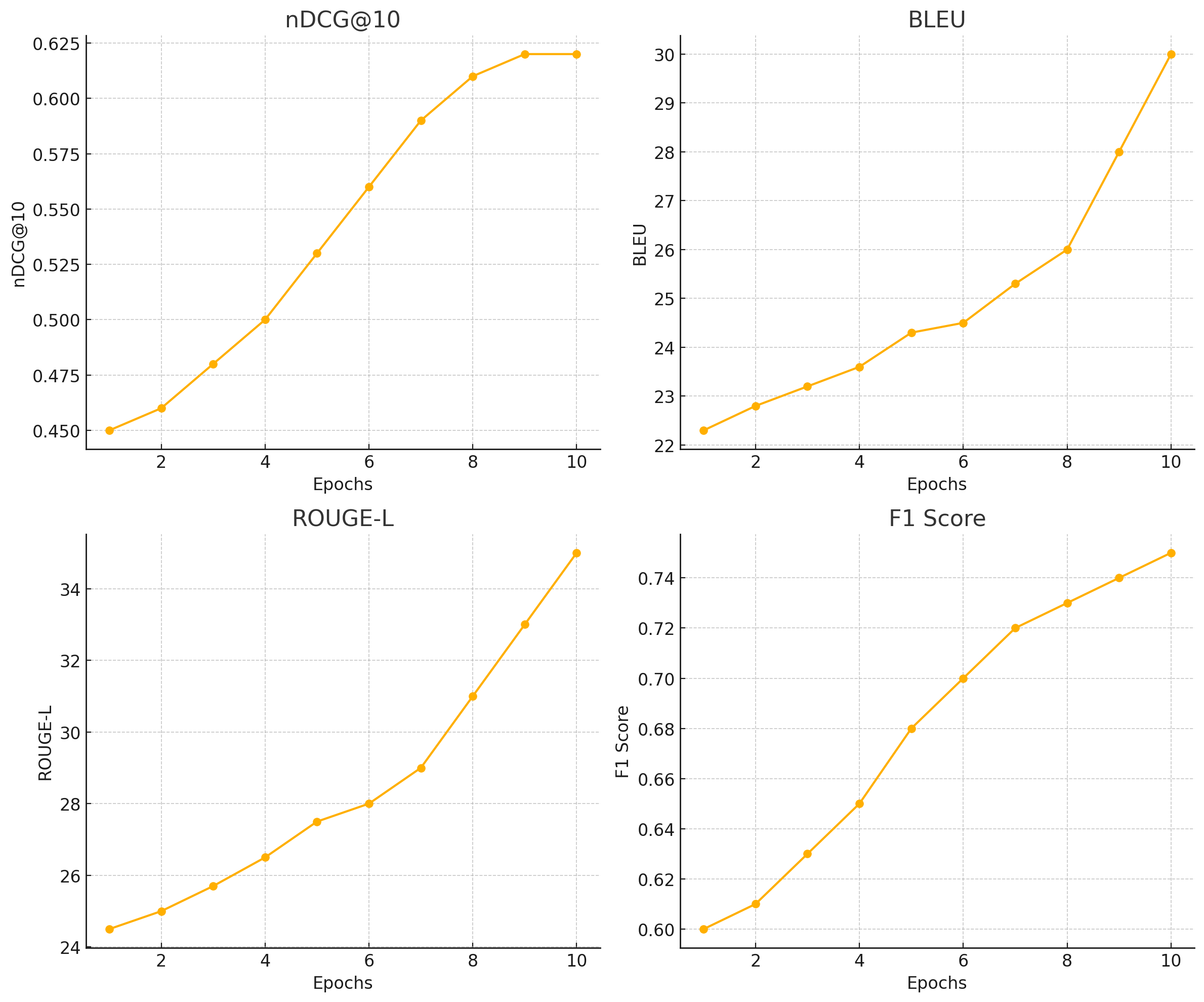}
  \caption{Changes in model training indicators over time.}
  \label{fig:metric2}
\end{figure}

Table~\ref{tab:full_evaluation} and Table~\ref{tab:ablation_study} summarize the performance of all models on five datasets using nDCG@10, BLEU, ROUGE-L, and F1 scores. Figure~\ref{fig:metric2} shows the changes in model training indicators.

\begin{table}[htbp]
  \caption{Full Model Evaluation Results}
  \label{tab:full_evaluation}
  \centering
  \resizebox{\columnwidth}{!}{%
    \begin{tabular}{|l|c|c|c|c|c|}
      \hline
      Model                   & FinDER (nDCG@10) & FinQABench (BLEU) & FinanceBench (ROUGE-L) & TATQA (F1) & FinQA (F1) \\ \hline
      BERT-based Retriever    & 0.45             & 22.3              & 24.5                   & 0.60       & 0.63       \\ \hline
      Traditional RAG         & 0.49             & 23.5              & 26.3                   & 0.62       & 0.67       \\ \hline
      FinBERT                 & 0.52             & 24.7              & 28.0                   & 0.64       & 0.70       \\ \hline
      GPT-3                   & 0.56             & 26.3              & 29.2                   & 0.66       & 0.72       \\ \hline
      FinLLaMA-RAG            & \textbf{0.62}    & \textbf{30.5}     & \textbf{35.2}          & \textbf{0.75} & \textbf{0.78} \\ \hline
      Retrieval-Only Model    & –                & –                 & –                      & –          & –          \\ \hline
      Generation-Only Model   & –                & –                 & –                      & –          & –          \\ \hline
    \end{tabular}%
  }
\end{table}

\begin{table}[htbp]
  \caption{Ablation Study Results}
  \label{tab:ablation_study}
  \centering
  \resizebox{\columnwidth}{!}{%
    \begin{tabular}{|l|c|c|c|c|}
      \hline
      Model                   & nDCG@10 & BLEU  & ROUGE-L & F1   \\ \hline
      BERT-based Retriever    & –       & –     & –       & –    \\ \hline
      Traditional RAG         & –       & –     & –       & –    \\ \hline
      FinBERT                 & –       & –     & –       & –    \\ \hline
      GPT-3                   & –       & –     & –       & –    \\ \hline
      FinLLaMA-RAG            & \textbf{0.62} & \textbf{30.5} & \textbf{35.2} & \textbf{0.75} \\ \hline
      Retrieval-Only Model    & 0.45    & 18.2  & 22.5    & 0.60 \\ \hline
      Generation-Only Model   & 0.48    & 19.1  & 24.1    & 0.62 \\ \hline
    \end{tabular}%
  }
\end{table}

\section{Conclusion}
In this paper, we introduced FinLLaMA-RAG, a novel Retrieval-Augmented Generation model for document analysis. Building on its strengths in complex financial QA, FinLLaMA-RAG also extends naturally into tax compliance and strategy—whether as a virtual tax assistant for individuals and SMEs or as a corporate tool for international transfer-pricing analysis. The model combines advanced retrieval techniques with a powerful generation model and multi-hop reasoning.

\bibliographystyle{IEEEtran} \bibliography{references}

\end{document}